# Rethinking the Image Feature Biases Exhibited by Deep CNN Models

Dawei Dai[1], Yutang Li[1], Huanan Bao[1], Sy Xia[1], Guoyin Wang[1], Xiaoli Ma[1]
1. *College of Computer Science and Technology*, *Chongqing University of Posts and Telecommunications*, *Chongqing*, China
dw_dai@163.com

## Abstract

In recent years, convolutional neural networks (CNNs) have been applied successfully in many fields. However, such deep neural models are still regarded as "black box" in most tasks. One of the fundamental issues underlying this problem is understanding which features are most influential in image recognition tasks and how they are processed by CNNs. It is widely accepted that CNN models combine low-level features to form complex shapes until the object can be readily classified, however, several recent studies have argued that texture features are more important than other features. In this paper, we assume that the importance of certain features varies depending on specific tasks, i.e., specific tasks exhibit a feature bias. We designed two classification tasks based on human intuition to train deep neural models to identify anticipated biases. We devised experiments comprising many tasks to test these biases for the ResNet and DenseNet models. From the results, we conclude that (1) the combined effect of certain features is typically far more influential than any single feature; (2) in different tasks, neural models can perform different biases, that is, we can design a specific task to make a neural model biased toward a specific anticipated feature.

## 1. INTRODUCTION

In recent years, following the breakthrough of convolutional neural network (CNN) models in the field of image processing, the application of CNNs has experienced explosive growth. CNN models have been deployed in a variety of applications, including automated vehicles, cancer detection, and recognition systems, reaching, and sometimes exceeding, the capabilities of human beings in many scenarios. Although CNN models perform impressively in many fields, only a few seemingly obscure model parameters and very high fitting evaluation results can be obtained. Although development in understanding the underlying mechanisms of such deep neural networks is expected to continue, the perception of current CNN models as "black box" structures persists. In fact, there is widespread concern regarding the application of deep CNN models, not only because the model itself cannot provide sufficient information, but also in terms of security. **However, considering the challenges highlighted here, the usefulness and security of these artificial intelligence (AI) systems will be limited by our ability to understand and control them.**

Deep neural network models are more difficult to understand or interpret than traditional approaches. Before CNN models were widely adopted, manner-designed operators were the standard approach in computer vision, for which the operator processes and the treatment of low-level features were well established. Examples of such methods include SIFT [Lowe, 2004] and HOG [Dalal et al., 2005]. For small datasets, traditional algorithms incorporating suitable operators perform very well. By contrast, for large datasets, the performance of traditional methods, such as ImageNet [Deng et al., 2009], is often significantly worse than deep CNN models [Krizhevsky et al., 2012]. The black-box characteristics of neural network models reflect two sides of the same coin. ***How are CNN models able to achieve impressive performance for complex tasks?*** It is widely accepted that CNN models combine low-level features to form increasingly complex shapes until the object can be readily classified; this term is referred to as "shape hypothesis". Some recent studies have suggested that image textures play a more important role than other features, that is, image processing contains a "texture bias"; however, a degree of controversy has arisen surrounding this issue.

Undeniably, texture and shape are both important features and play important roles in image recognition tasks. **However, we consider which a neural model bias toward a particular feature depends on the nature of the task**. Here, the "task" refers specifically to a combination of input data. Taking image recognition as an example, the classification of categories A and B, and the classification of categories A and C are considered as two different "tasks" (task1 and task2, respectively); category A is a common category shared by the two tasks, whereas we design the other categories carefully to ensure that the main distinguishing features for category A are different for each task. We perform extensive experiments using the standard architectures of the ResNet and DenseNet models for the designed tasks. From the experiments, we note that (1) feature diversity is a greater factor in image recognition tasks than any single feature by itself; (2) for the same object recognition, shape and texture information can be both extremely important or inconsequential depending on the task; (3) shape can be more important than texture, and vice versa. Therefore, we propose that the so-called bias toward certain features is an artifact of neural model performance for the particular task.

## 2. Related works

### 2.1 Shape hypothesis

From the perspective of cognitive psychology, people are very sensitive to shape [Ritteret al., 2017]. Deep CNN models were widely considered to combine low-level features layer by layer to construct shapes until the objects could be readily classified. As Kriegeskorte stated "neural network model acquires complex knowledge about the kinds of shapes associated with each category" [Kriegeskorte, 2015]. This concept has also been expressed by LeCun: "intermediate layers in a CNN model recognize parts of familiar objects and subsequent layers can detect objects as combinations of the parts" [LeCun, 2015].

The shape hypothesis was also supported by a number of empirical findings. For example, Kubilius et al. built CNN-based computational models for human shape perception and concluded that CNNs learn shape representations that reflect human shape perception implicitly [Kubilius et al, 2016]. "Shape bias" was also discovered in CNN models for image classification, that is, the shape features of an object can be more important than other features in object classification tasks [Ritter et al, 2017]. It has also been reported that some algorithms can extract object edges using richer convolutional features [Liu et al., 2017], thereby proving that CNNs have learned to identify edge features. Moreover, tasks such as human detection [Dalal et al., 2005] also use richer edge information. Visualization methods such as deconvolutional networks [Zeiler & Fergus, 2014] often highlight the object parts in high layers of CNNs. This was used to argue that shape plays a greater role in CNN models than other features.

## 2.2 Texture bias

Other studies have provided contrasting evidence that suggests that texture features are more important. This view originated with the bag-of-words (BOW) model [Li et al., 2009], which focuses solely on local features and abandons the global image structure. In this method, local features such as textures are more effective than global features such as shape. Following the increasing application of deep CNN models to image processing tasks, several studies have noted that deep CNN models can perform well even if the image topology is disrupted [Gatys et al., 2015], whereas standard CNN models, in which shapes are preserved yet texture cues are removed, perform worse for object sketch recognition [Ballester & de Araujo, 2016]. Other research has indicated that Imagenet-trained CNN models are biased toward texture and that increasing shape bias improves the accuracy and robusticity of image processing tasks [Geirhos et al., 2018]. Additionally, Gatys et al. [Gatys et al., 2015] suggested that local information such as textures are sufficient to "solve" ImageNet object recognition problems. Furthermore, using the style transfer algorithm, Robert et al. proved that texture features were more important for the ResNet-50 model than other features [Robert et al., 2018].

## 2.3 Summary

Of course, numerous other features can be utilized in image recognition. In this paper, our discussion focuses on the influence of texture and shape on the image recognition task. However, research into which is more important for image recognition tasks has generated some controversy. Following careful analysis of the aforementioned related studies, we find that the evaluative measures used to form conclusions are often different for individual studies. In short, CNN models may exhibit bias toward shape in certain tasks but may be biased toward texture for other tasks.

For a specific **task**, in fact, its real knowledge or distribution exists, but it is unknown to us. CNNs provide a means to learn or approximate this knowledge. **Therefore, we examine the extent to which a neural model exhibits bias toward a certain feature depending on the specific task. We design particular tasks to guide the deep neural models bias towards specific features based on our predictions.** The feasibility of this approach is verified through experiments, with the results implying that the so-called bias toward shape or texture features is linked inextricably to the CNN model performance for specific tasks.

# 3. Approach

## 3.1 Measurement of feature importance

In an image recognition task, the importance of a certain type of feature can be measured by its impact on recognition accuracy. As a result, the key to the experiment is to measure the drop in accuracy that results when a specific feature is removed or constrained. The more the accuracy decreases in response to removing or weakening a specific feature, the more important this feature is, thus revealing the extent to which a CNN model is biased toward the important features. Here, we use the feature contribution rate (FCR) to describe this process. As expressed in Eq. 1, the ***$acc_o$*** term indicates the classification accuracy of the original images, whereas ***$acc_{rm}$*** indicates the accuracy for images in which some specific features have been removed or weakened. Thus, the smaller the value of ***FCR***, the more important the feature is, and vice versa:

$$FCR = 1 - \frac{acc_o - acc_{rm}}{acc_o}. \quad (1)$$

## 3.2 Removing (or weakening) certain features

The current debate focuses on which type of feature exerts greater influence within CNN models, texture, shape, or others? Therefore, we analyzed the respective influence or importance of texture and shape in an image recognition task. Practically, it is difficult to remove a certain type of feature completely, in which case, the relevant feature was suppressed rather than eliminated.

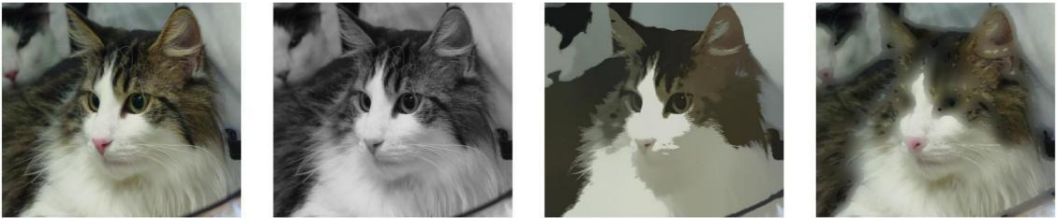

Figure. 1 Images for which a certain feature has been removed or weakened: (a) original image; (b) image with the colors removed; (c) image with the texture weakened; (d) image with the shape edges weakened.

**Color removal.** Color originates from the superposition of different wavelengths of light. For RGB pictures, the color feature is defined by the difference between R, G, and B channels. We used the brightness information of the image, which is the average of the R, G, and B channels, to replace the R, G, and B channels (see Fig. 1(b)).

**Texture weakening.** Texture refers to the pixels of the picture that obey certain statistical properties. In the frequency domain, textures are usually concentrated in a certain frequency band. We used the MeanShiftFilter [Fukunaga & Hostetler, 1975] method to remove texture information without destroying edges. The resulting image is shown in Fig. 1(c).

**Shape weakening.** Shape refers to a region that is described by some low-level features. We weakened the shape by applying edge blurring. The edge extraction algorithm was first used to detect boundaries [Leordeanu et al., 2012], then a Gaussian filter was used to blur the edges. The result is shown in Fig. 1(d).

**Destroying topology.** Topology features are very important for human identification. For example, a clock must have a dial and hands. Individually, the dial and hands are not considered to be a clock. To implement destruction of topology, we split the picture and shuffle the position of each piece, analogous to a Rubik's cube.

### 3.3 Designing the tasks

**Task 1: Diverse**
Neural models are required to learn different knowledge for different tasks. First, we consider that a deep neural network model may exhibit bias toward a diversity of features depending on the nature of the task. For a specific task, the "knowledge" was fixed; however, different models may learn this knowledge to differing degrees. As a result, we began by selecting several different tasks, with each task containing different "knowledge." Then, we trained several standard deep neural models via conventional methods and tested the influence of different features on different tasks. Last, we analyzed the results.

**Task 2: Importance of texture for the same object can be significantly different**
The data-driven learning of a neural model is designed so that the differences in an image recognition task are learned. In this case, **we designed a multi-classification image recognition task, in which the distinctive features between each category are known.** However, it is not easy to isolate a specific feature from others. Therefore, to simplify the task, we explored the importance of each category separately. We designed two subtasks (two datasets), with subtask1 containing category A, category B, and category C,... and subtask2 containing category A, category B1, and category C1,.... Both subtasks contain category A; in subtask1, only category A contains rich textures, whereas in subtask2, all categories contain rich textures. We explored the importance of texture in category A for the two subtasks. Intuitively, we can hypothesize that the textures of category A should be more important in subtask1 than that in subtask2.

**Task 3: Importance of shape for the same object can be significantly different**
We adopted a similar design to task2. Again, we began by designing two subtasks, each containing one common category, i.e., category A. There were two governing conditions for this task: (1) in subtask1, the shape of the object needed to be similar in all categories; (2) in subtask2, the shape of the object belonging to category A needed to be obviously different from the objects belonging to the other categories.

**Task 4: Texture (Shape) is more important than shape (Texture)**
In this paper, we need to design tasks on the basis that a deep neural model learns the differences between categories in image recognition tasks. Therefore, we carefully devise a subtask1 consisting of categories (A, B, C, ...), in which the objects in each category has a similar shape but their textures are obviously different. Conversely, we then designed subtask2, comprising categories (A, B1, C1, ...), in which the object texture is similar across all categories but the object shapes are obviously different. Intuitively, if a neural network performs well for both subtasks, this would indicate a bias toward texture in subtask1 and shape in subtask2.

### 3.4 Implementation

We choose 4 categories from ImageNet that contains 1000 categories to design each specific task. Each task (sub dataset) consisted of 1,500 training images and 50 validation images. We resize all images size to [224, 224] and the horizontal flip and image shift augmentation were in use. Thereafter, dividing images by 255 to provide them in range [0,1] as input.

In addition, we selected trained several standard neural networks: VGG [Simonyan et al.,2014], ResNet [He et al., 2016] (See Table I), and DenseNet [Huang et al., 2017] (See Table II) to explore the feature biases. We used weight decay of 1e-4 for all models, adam for ResNet and momentum of 0.9 for Densenet, and adopted the weight initialization and BN introduced by [He et al., 2016; Huang et al., 2016]. We trained all the models with a mini-batch size of 32 or 64 on 2080Ti GPU.

Table I Architecture for ResNet models

| Layer | Out size | Architecture |
|---|---|---|
| conv1 | 112 × 112 | 7 × 7, 4, stride 2 |
| Conv2 | 56 × 56 | 5 × 5, 4, stride 2 |
| Residual block layers 1 | 56 × 56 | $\begin{bmatrix} 3 \times 3, 4 \\ 3 \times 3, 4 \end{bmatrix} \times 2$ |
| Residual block layers 2 | 28 × 28 | $\begin{bmatrix} 3 \times 3, 8 \\ 3 \times 3, 8 \end{bmatrix} \times 2$ |
| Residual block layers 3 | 14 × 14 | $\begin{bmatrix} 3 \times 3, 16 \\ 3 \times 3, 16 \end{bmatrix} \times 2$ |
| Residual block layers 4 | 7 × 7 | $\begin{bmatrix} 3 \times 3, 32 \\ 3 \times 3, 32 \end{bmatrix} \times 2$ |
| Classification Layer | 1 × 1 | global average pool |
| | | 2-d fc, softmax |

Table II Architecture for DenseNet models

| Layer | Out size | Architecture |
|---|---|---|
| conv1 | 112 × 112 × 24 | 7 × 7, 24, stride 2 |
| Dense block layers 1 | 56 × 56 × 48 | 3 × 3 max pool, stride 2 |
| | | $\begin{bmatrix} 1 \times 1, 48 \\ 3 \times 3, 12 \end{bmatrix} \times 3$ |
| Transition Layer 1 | 56 × 56 × 24 | 1 × 1 conv |
| | 28 × 28 × 24 | 2×2 max pool, stride 2 |
| Dense block layers 2 | 28 × 28 × 48 | $\begin{bmatrix} 1 \times 1, 48 \\ 3 \times 3, 12 \end{bmatrix} \times 3$ |
| Transition Layer 2 | 28 × 28 × 24 | 1 × 1 conv |
| | 14 × 14 × 24 | 2×2 max pool, stride 2 |
| Dense block layers 3 | 14 × 14 × 48 | $\begin{bmatrix} 1 \times 1, 48 \\ 3 \times 3, 12 \end{bmatrix} \times 3$ |
| Transition Layer 3 | 14 × 14 × 24 | 1×1 conv |
| | 7 × 7 × 24 | 2×2 max pool, stride 2 |
| Dense block layers 4 | 7 × 7 × 48 | $\begin{bmatrix} 1 \times 1, 48 \\ 3 \times 3, 12 \end{bmatrix} \times 3$ |
| Classification Layer | 1 × 1 × 48 | global average pool |
| | | 2-d fc, softmax |

## 4. Experiments and Analyses

### 4.1 Tendency of neural network model is diverse

We selected four types of features (color, texture, edge, and image topology) to discuss their importance to deep neural networks in image recognition tasks. Additionally, we selected four models (VGG, ResNet, Inception, DenseNet) and trained them using the ImageNet dataset. Then, we selected 100 classes from the training dataset at random and removed each feature type systematically to assess their individual contributions. Finally, we tested the performance of the neural models for these datasets.

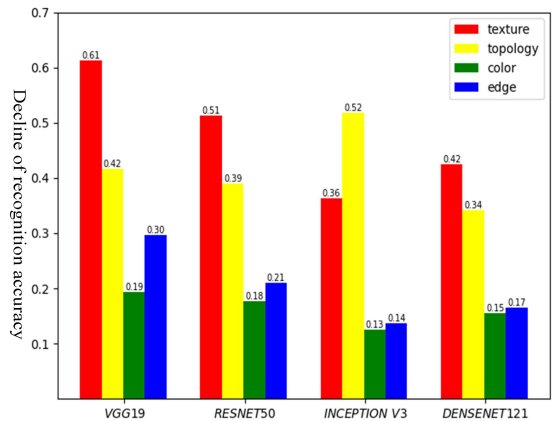

Figure. 2 Contributions of the different feature types for each CNN.

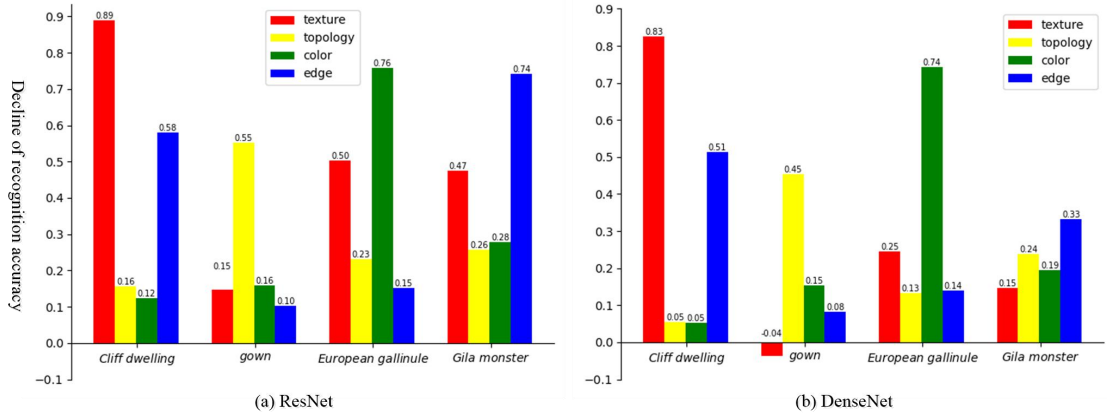

Figure. 3 Contributions of the different feature types to selected categories for the ResNet and DenseNet.

The average performance of the 100-classes datasets is shown in Fig. 2. We note the following observations: **(1)** for a specific task, the importance of different features for different neural models can be similar; **(2)** in this task, texture and topology features were more important than color and edges. As shown in Fig. 3, we tested the importance of four features in four categories, revealing that the most important feature of deep models was different for each category, with each type of feature exerting greater influence for different categories in different tasks. Consequently, we note that the observed biases were not unique to specific classes. Equipped with this information, we attempted to design a task enabling the deep neural models to exhibit bias toward the specific features based on our predictions.

### 4.2 Texture importance analysis

**A. Task design**

We designed two tasks (see Fig. 4), each containing four categories: task1: **pig**, cat, dog, and lion; task2: **pig,** hippopotamus, walrus, and sea lion. For task1, all four animals are rich in hair, representing similarity with respect to their texture features, whereas for task2, only the category "**pig**" is rich in hair. Intuitively, in task1, it is not easy to identify the **pig** using local texture features only. By contrast, in task2, it is relatively easy to identify the "pig" by the local texture. Although, we cannot avoid the influence of other features completely, we just minimize their impact. We can analyze the importance of texture for the category "**pig**" via the two tasks, and thus verify whether the neural network model conforms to our intuition, as follows:

**Step1**: We trained several standard deep neural models, including the ResNet and DenseNet models, to perform task1 and task2.

**Step2**: We weakened the texture of the "**pig**" category in two tasks (see Fig. 5).

**Step3**: We tested the performance of the deep neural models with the weakened texture "**pig**" category, and analyze the importance of texture for the "**pig**" category in different tasks.

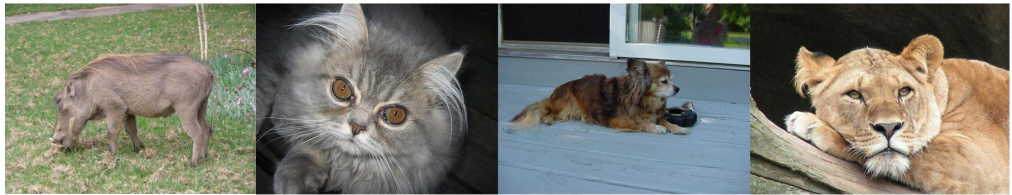

(a) Task1: **Pig**, Cat, Dog, Lion.

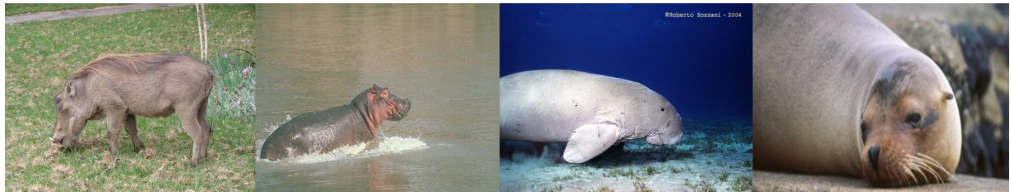

(b) Task2: **Pig**, Hippopotamus, Walrus, Sea lion.

Figure. 4 Images used in the two tasks analyzing the importance of texture.

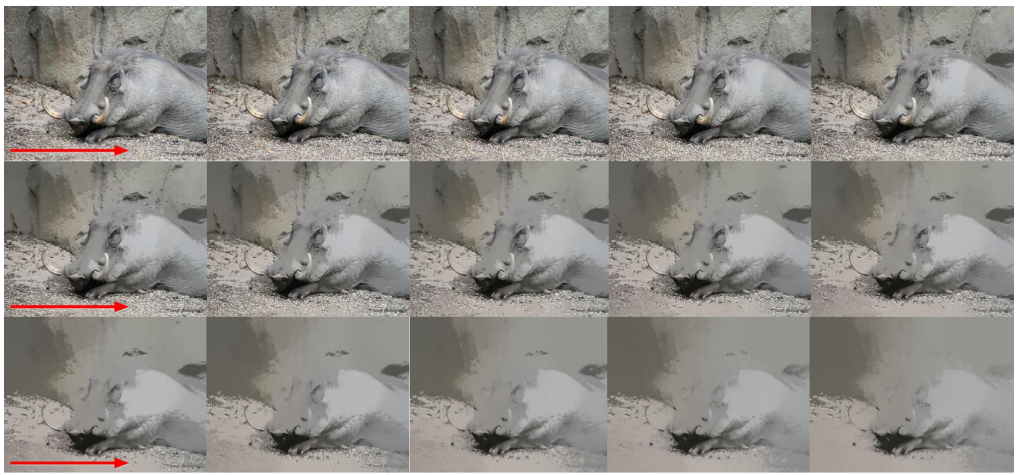

Figure. 5 Images illustrating the weakening of texture features. The texture becomes successively weaker from left to right in each row.

## B. Results

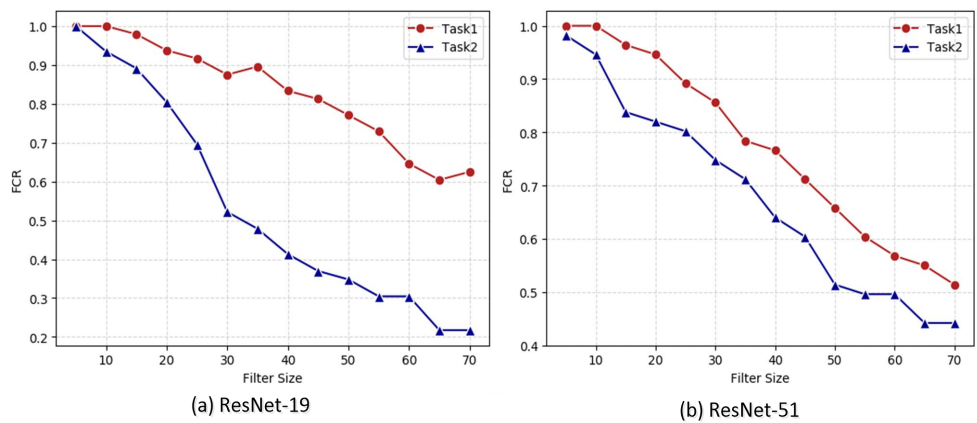

(a) ResNet-19

(b) ResNet-51

**Figure. 6** Texture importance in the ResNet model for each task. The horizontal axis represents the size of the filter window, which is positively correlated with weakening strength. (a) ResNet-19; (b) ResNet-51.

As shown in Figs. 6 7, the recognition ability of both the ResNet and DenseNet models for the "**pig**" category with weakened texture in task2 decreased more than in task1. It should be noted that the texture of "**pig**" in task2 is much more important than in task1. In task2, texture features play an important role in both ResNet and DenseNet, whereas in task1, especially for the DenseNet model, the texture feature is less influential. Therefore, the importance of the same feature of the same category can differ significantly between the two tasks.

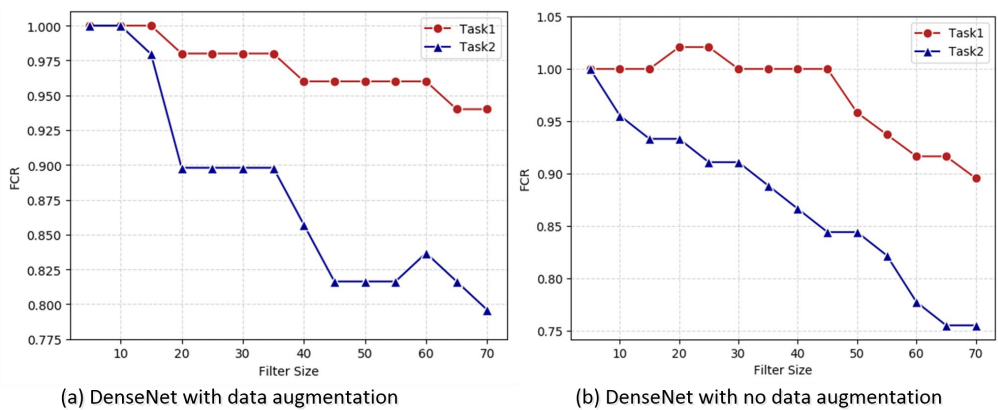

(a) DenseNet with data augmentation        (b) DenseNet with no data augmentation

**Figure. 7** Texture importance in the DenseNet model for each task: (a) Without data augmentation; (b) With data augmentation.

### 4.3 Shape importance analysis

**A. Task design**

As in Section 4.2, two tasks were designed to analyze the importance of shape features, as shown in **Fig. 8:** task1: **fox**, husky, wolf, golden retriever; task2: **fox, rabbit,** horse, bear For task1, all four animals have similar shapes, whereas in task2, the shape of the category "**fox**" was obviously different from the other three. As such, in task1, it is not easy to identify the **fox** based on shape features alone, whereas it is relatively easy to identify the fox using shape features in task2. We analyzed the importance of the shape of category "**fox**" in each task as follows:

   **Step1**: We trained several standard deep neural models to perform task1 and task2.

   **Step2**: We weakened the shape of the "**fox**" category in each task (see Fig. 9).

   **Step3**: We tested the performance of the deep neural models with the weakened shape of the "**fox**" category, and analyzed the importance of shape for the "**fox**" category in different tasks.

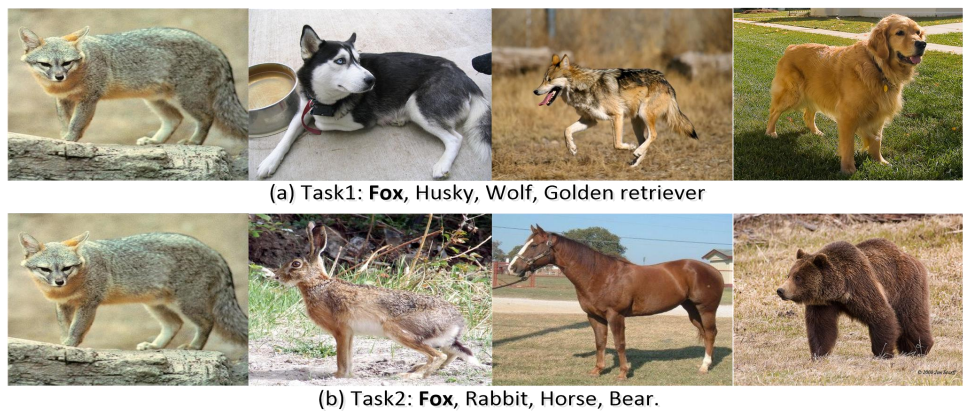

(a) Task1: **Fox**, Husky, Wolf, Golden retriever

(b) Task2: **Fox**, Rabbit, Horse, Bear.

**Figure. 8** Images used in the two tasks for analyzing the importance of shape.

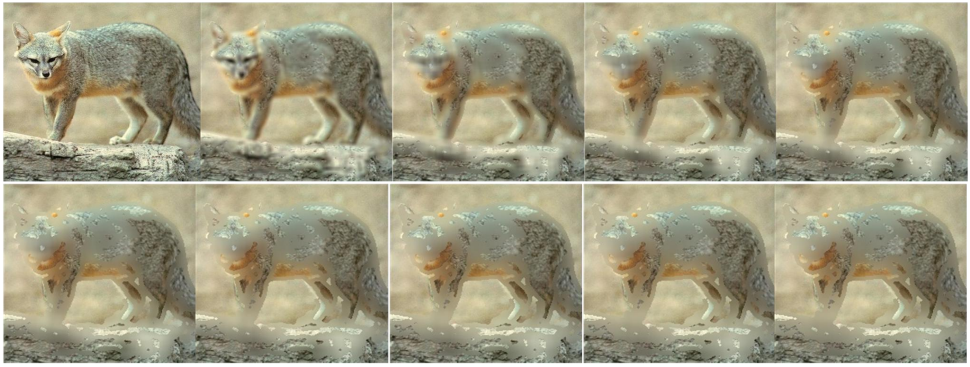

**Figure. 9** Images illustrating the weakening of shape features. The shape becomes successively weaker from left to right in each row.

### B. Results

As shown in Fig. 10, the recognition ability of ResNet and DenseNet models with respect to the weakened shape of the category "**fox**" decreased more for task2 than for task1. It should be noted that the shape features of the "**fox**" category are far more important in task2 than in task1.

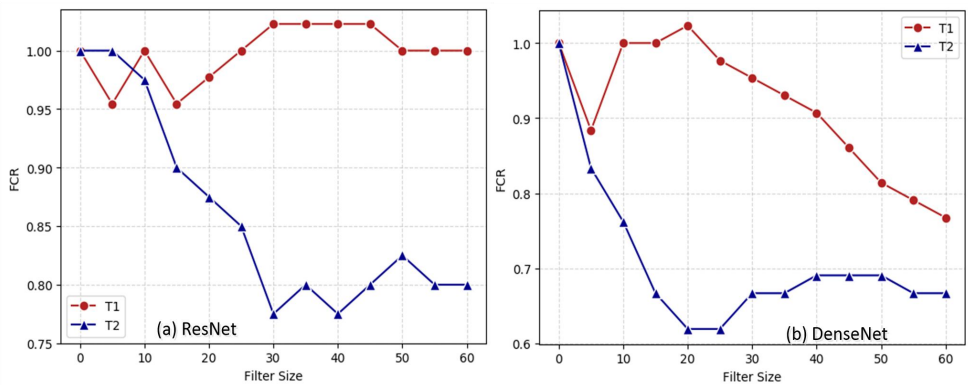

**Figure. 10** Shape importance in the ResNet and DenseNet models for each task.

### 4.4 Which is more important, texture or shape?

**A. Task design**

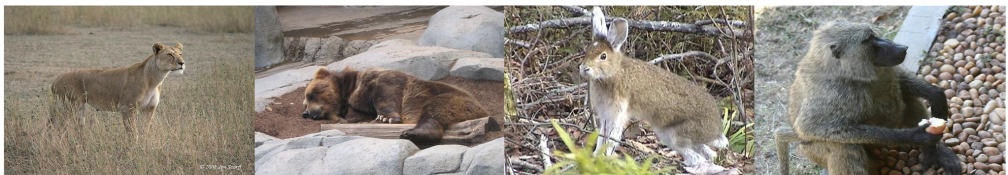
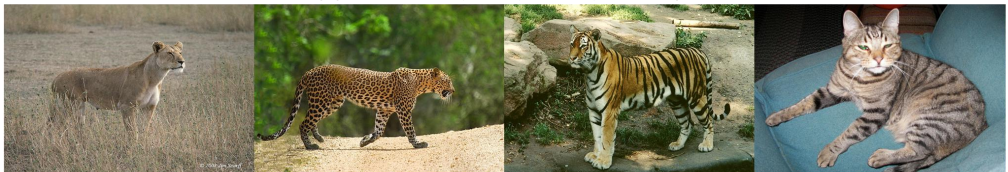

**Figure. 11** Images used in each task for the comparative analysis of shape and texture importance.

As shown in Fig. 11, we designed two further tasks: task1: **lion,** bear, rabbit, monkey; task2: **lion**, cheetah, tiger, cat. In task1, the shape features are significantly different between the lion and the other three animals, whereas in task2, the texture features represent the primary differentiator between the lion and the other categories. We consider that deep models learn these differences exactly so that the model can be applied to image recognition tasks. We evaluate and compare the respective importance of texture and shape, as follows:

**Step1**: We trained several standard deep neural models to perform task1 and task2.

**Step2**: We obtained two groups of modified data: one in which the shape features of "**lion**" were weakened compared to the original image, and one in which the texture features of "**lion**" were weakened relative to the original image. The corresponding images are presented in Fig. 12.

**Step3**: We tested the performance of the deep neural models on modified category "lion", to determine which is more important, shape or texture.

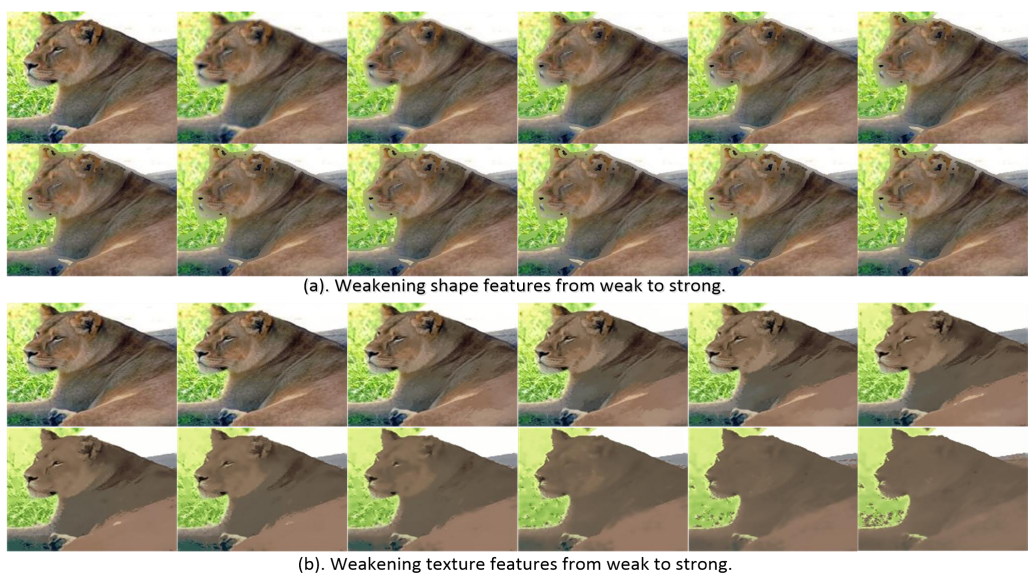

(a). Weakening shape features from weak to strong.

(b). Weakening texture features from weak to strong.

**Figure. 12** Images illustrating the weakening applied to the lion category image with respect to (a) shape and (b) texture. The amount of weakening increases from left to right in each row of images.

**B. Results**

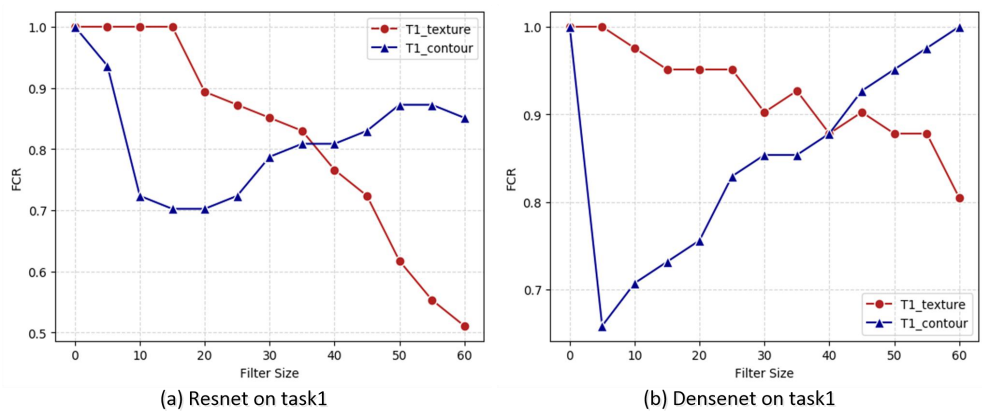

(a) Resnet on task1

(b) Densenet on task1

**Figure. 13** Comparison of shape and texture importance for the ResNet and DenseNet models for task1.

The results of the comparative analysis are illustrated in Fig. 13. In task1, shape information can be significantly more important than that of texture when considering slightly weakened edge for image recognition performed by both the ResNet and DenseNet models. Conversely, when the edge is heavily weakened, the reverse behavior can be observed. Ultimately, it is extremely challenging to remove the shape of an object completely via edge weakening; weakening the outline of the object only creates new edges, as shown in Fig. 14. This explains the initial decrease in the performance of the models, with the trend reversing as the filter size increases (blue lines in Fig. 13). As shown in Fig. 15, for task2, we can conclude that the texture information contributes more than the shape information.

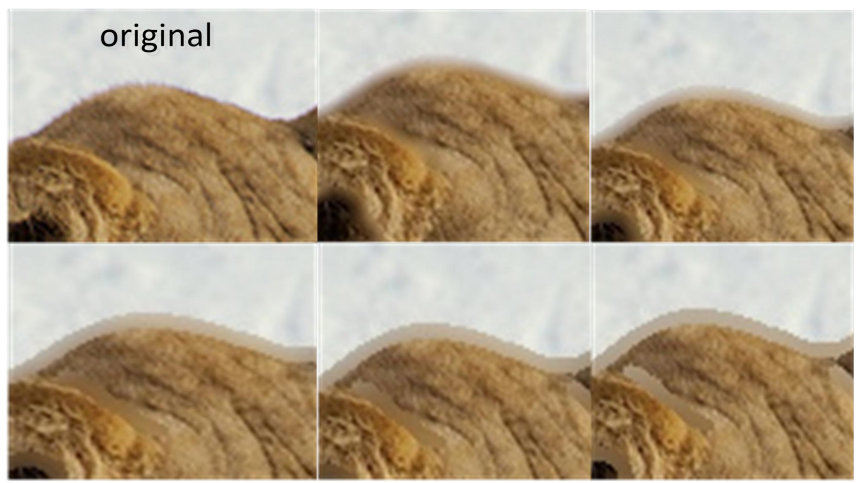

**Figure. 14** Applying increasing degrees of contour weakening to the original image (a) leads to the emergence of an increasing number of new edges.

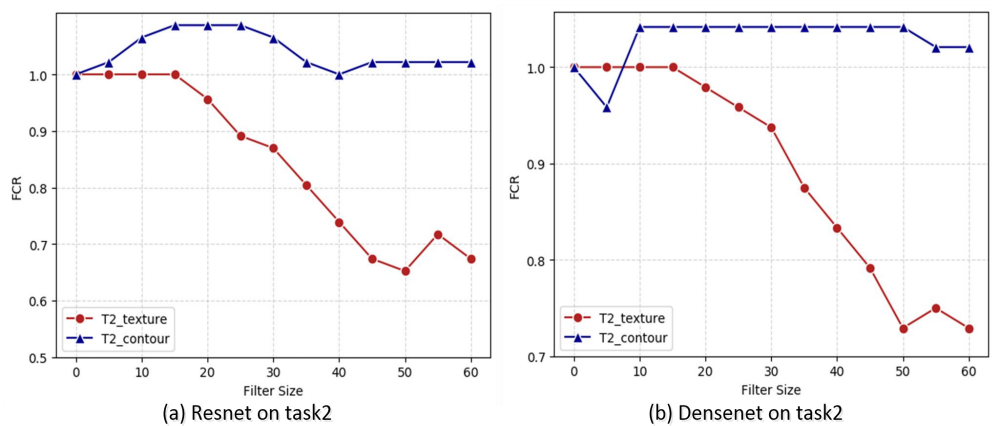

**Figure. 15** Comparison of shape and texture importance for the ResNet and DenseNet models for task2.

# 5. Conclusion

Although CNNs have been applied successfully to a variety of AI systems, deep neural models are regarded as "black boxes" in many tasks. This perception arises from the fact that the influence of features in image recognition tasks remains largely unknown. It has been posited that CNN models combine low-level features to form complex shapes until the object can be readily classified; however, recent contrasting studies have claimed that texture features are the real key to effective image recognition.

We propose that these reported biases are task specific, as clear image feature bias disparities can be observed for different tasks. In our experiments, which consider the same "object" in different tasks, many outcomes are possible; for example, texture (shape) can play an important role but, equally, can also be insignificant. In certain cases texture features are more important than shape features and vice versa. Therefore, such biases can be designed manually, that is, we can design a particular task to create a deep neural model bias toward a specific type of feature that we consider vitally important to a particular task.

# References

[Ballester & de Araujo, 2016] Pedro Ballester and Ricardo Matsumura de Araujo. On the performance of GoogLeNet and AlexNet applied to sketches. In AAAI, pp. 1124–1128, 2016.

[Dalal et al., 2005] Dalal N, Triggs B. Histograms of oriented gradients for human detection[C]. 2005.

[Deng et al., 2009] Deng J, Dong W, Socher R. Imagenet: A large-scale hierarchical image database[C] //2009 IEEE conference on computer vision and pattern recognition. Ieee, 2009: 248-255.

[Fukunaga & Hostetler, 1975] Fukunaga, K. & Hostetler, L. The estimation of the gradient of a density function. IEEE Transactions on Information Theory - TIT. 1975.

[Geirhos et al., 2018] Geirhos R , Rubisch P , Michaelis C , et al. ImageNet-trained CNNs are biased towards texture; increasing shape bias improves accuracy and robustness[J]. 2018.

[Gatys et al., 2015] Leon A Gatys, Alexander S Ecker, and Matthias Bethge. Texture synthesis using convolutional neural networks. In Advances in Neural Information Processing Systems, pp. 262–270, 2015.

[He et al., 2016] He K , Zhang X , Ren S , et al. Deep Residual Learning for Image Recognition[J]. 2016.

[Huang et al., 2016] Huang G , Liu Z , Laurens V D M , et al. Densely Connected Convolutional Networks[J]. 2016.

[Krizhevsky et al., 2012] Krizhevsky A, Sutskever I, Hinton G E. Imagenet classification with deep convolutional neural networks[C] //Advances in neural information processing systems. 2012: 1097 - 1105.

[Kriegeskorte, 2015] N. Kriegeskorte. Deep neural networks: A new framework for modeling biological vision and brain information processing. Annual Review of Vision Science, 1(15):417–446, 2015.

[Kubilius et al, 2016] Jonas Kubilius, Stefania Bracci, and Hans P Op de Beeck. Deep neural networks as a computational model for human shape sensitivity. PLoS Computational Biology, 12(4):e1004896, 2016.

[Lowe, 2004] Lowe D G. Distinctive image features from scale-invariant keypoints[J]. International journal of computer vision, 2004, 60(2): 91-110.

[Leordeanu et al., 2012] Leordeanu M, Sukthankar R, Sminchisescu C. Efficient closed-form solution to generalized boundary detection[C] //European Conference on Computer Vision. Springer, Berlin, Heidelberg, 2012: 516-529.

[LeCun et al., 2015] Y. LeCun, Y. Bengio, and G. Hinton. Deep learning. Nature, 521(7553):436–444, 2015.

[Liu et al., 2017] Liu Y, Cheng M M, Hu X, et al. Richer convolutional features for edge detection[C] // Proceedings of the IEEE conference on computer vision and pattern recognition. 2017: 3000-3009.

[Li et al., 2009] Li Z, Imai J, Kaneko M. Facial-component-based bag of words and phog descriptor for facial expression recognition[C] // 2009 IEEE International Conference on Systems, Man and Cybernetics. IEEE, 2009: 1353-1358.

[Ritter et al., 2017] Ritter S, Barrett D G T, Santoro A, et al. Cognitive psychology for deep neural networks: A shape bias case study[C]//Proceedings of the 34th International Conference on Machine Learning-Volume 70. JMLR. org, 2017: 2940-2949.

[Robert et al., 2018] Robert Geirhos, Carlos M. Medina Temme, Jonas Rauber, Heiko H Schutt, Matthias Bethge, and Felix A Wichmann. Generalisation in humans and deep neural networks. arXiv preprint arXiv: 1808.08750, 2018.

[Simonyan et al., 2014] Simonyan, Karen, and Andrew Zisserman. "Very deep convolutional networks for large-scale image recognition." arXiv preprint arXiv:1409.1556 (2014).

[Zeiler & Fergus, 2014] Matthew D Zeiler and Rob Fergus. Visualizing and understanding convolutional networks. In European Conference on Computer Vision, pp. 818–833. Springer, 2014.